# LifeCLEF Plant Identification Task 2014

Hervé Goëau[1], Alexis Joly[1,2], Pierre Bonnet[3], Souheil Selmi[1], Jean-François Molino[4], Daniel Barthélémy[5], and Nozha Boujemaa[6]

[1] Inria ZENITH team, France, `name.surname@inria.fr`
[2] LIRMM, Montpellier, France
[3] CIRAD, UMR AMAP, France, `pierre.bonnet@cirad.fr`
[4] IRD, UMR AMAP, France, `wjean-francois.molino@ird.fr`
[5] CIRAD, BIOS Direction and INRA, UMR AMAP, F-34398, France
`daniel.barthelemy@cirad.fr`
[6] INRIA, Direction of Saclay Center, `nozha.boujemaa@inria.fr`

**Abstract.** The LifeCLEFs plant identification task provides a testbed for a system-oriented evaluation of plant identification about 500 species trees and herbaceous plants. Seven types of image content are considered: scan and scan-like pictures of leaf, and 6 kinds of detailed views with unconstrained conditions, directly photographed on the plant: flower, fruit, stem & bark, branch, leaf and entire view. The main originality of this data is that it was specifically built through a citizen sciences initiative conducted by Tela Botanica, a French social network of amateur and expert botanists. This makes the task closer to the conditions of a real-world application. This overview presents more precisely the resources and assessments of task, summarizes the retrieval approaches employed by the participating groups, and provides an analysis of the main evaluation results. With a total of ten groups from six countries and with a total of twenty seven submitted runs, involving distinct and original methods, this fourth year task confirms Image & Multimedia Retrieval community interest for biodiversity and botany, and highlights further challenging studies in plant identification.

## 1 Introduction

Content-based image retrieval approaches are nowadays considered to be one of the most promising solution to help bridge the botanical taxonomic gap, as discussed in [10] or [22] for instance. We therefore see an increasing interest in this trans-disciplinary challenge in the multimedia community (e.g. in [12], [5], [20], [23], [14], [2]). Beyond the raw identification performances achievable by state- of-the-art computer vision algorithms, the visual search approach offers much more efficient and interactive ways of browsing large floras than standard field guides or online web catalogs. Smartphone applications relying on such image- based identification services are particularly promising for setting-up massive ecological monitoring systems, involving hundreds of thousands of contributors at a very low cost.

Noticeable progress in this way was achieved by several project and apps like LeafSnap[7] [22], PlantNet[8],[9] [16]. But as promising as these applications are, their performances are however still far from the requirements of a real-world social-based ecological surveillance scenario. Allowing the mass of citizens to produce accurate plant observations requires to equip them with much more accurate identification tools. Measuring and boosting the performances of content-based identification tools is therefore crucial. This was precisely the goal of the ImageCLEF[10] plant identification task organized since 2011 in the context of the worldwide evaluation forum CLEF[11](see [10], [11] and [17] for more details).

Contrary to previous evaluations reported in the literature, the key objective was since the first campaign to build a realistic task closer to real-world conditions (different users, cameras, areas, periods of the year, individual plants, etc.). This was initially achieved through a citizen science initiative initiated 4 years ago in the context of the Pl@ntNet project in order to boost the image production of Tela Botanica social network. The evaluation data was enriched each year with the new contributions and progressively diversified with other input feeds (annotation and cleaning of older data, contributions made through Pl@ntNet mobile applications). The plant task of LifeCLEF 2014 is directly in the continuity of this effort. Main novelties compared to the last years are the following:

- an explicit multi-image query scenario,
- the supply of user ratings on image quality in the meta-data,
- a new type of view called "Branch" additionally to the 6 previous ones,
- and basically more species: 500 which is an important step towards covering the entire flora of a given region.

## 2   Dataset

More precisely, PlantCLEF 2014 dataset is composed of 60,962 pictures belonging to 19,504 observations of 500 species of trees, herbs and ferns living in a European region centered around France. This data was collected by 1608 distinct contributors. Each picture belongs to one and only one of the 7 types of view reported in the meta-data (entire plant, fruit, leaf, flower, stem, branch, leaf scan) and is associated with a single plant observation identifier allowing to link it with the other pictures of the same individual plant (observed the same day by the same person). It is noticeable that most image-based identification methods and evaluation data proposed in the past were so far based on leaf images (e.g. in [22], [3], [5] or in the more recent methods evaluated in [11]). Only few of them were focused on flower's images as in [25] or [1]. Leaves are far from being the only discriminant visual key between species but, due to their shape and size, they have the advantage to be easily observed, captured and described. More diverse parts of the plants however have to be considered for accurate identification.

---

[7] http://leafsnap.com/
[8] https://play.google.com/store/apps/details?id=org.plantnet&hl=en
[9]  http://identify.plantnet-project.org/
[10] http://www.imageclef.org/
[11] http://www.clef-initiative.eu/

An originality of PlantCLEF dataset is that its social nature makes it closer to the conditions of a real-world identification scenario: (i) images of the same species are coming from distinct plants living in distinct areas (ii) pictures are taken by different users that might not used the same protocol to acquire the images (iii) pictures are taken at different periods in the year. Each image of the dataset is associated with contextual meta-data (author, date, locality name, plant id) and social data (user ratings on image quality, collaboratively validated taxon names, vernacular names) provided in a structured xml file. The gps geo-localization and the device settings are available only for some of the images.

Table 1 gives some examples of pictures with decreasing averaged users ratings for the different types of views. Note that the users of the specialized social network creating these ratings (Tela Botanica) are explicitly asked to rate the images according to their plant identification ability and their accordance to the pre-defined acquisition protocol for each view type. This is not an aesthetic or general interest judgement as in most social image sharing sites.

To sum up each image is associated with the following meta-data:

- **ObservationId**: the plant observation ID from which several pictures can be associated
- **FileName**
- **MediaId**: id of the image
- View **Content**: Branch or Entire or Flower or Fruit or Leaf or LeafScan or Stem
- **ClassId**: the class number ID that must be used as ground-truth. It is a numerical taxonomical number used by Tela Botanica
- **Species** the species names (containing 3 parts: the Genus name, the Species name, the author(s) who discovered or revised the name of the species)
- **Genus**: the name of the Genus, one level above the Species in the taxonomical hierarchy used by Tela Botanica
- **Family**: the name of the Family, two levels above the Species in the taxonomical hierarchy used by Tela Botanica
- **Date**: (if available) the date when the plant was observed,
- **Vote**: the (round up) average of the user ratings on image quality
- **Location**: (if available) locality name, most of the time a town
- **Latitude & Longitude**: (if available) the GPS coordinates of the observation in the EXIF metadata, or, if no GPS information were found in the EXIF, the GPS coordinates of the locality where the plant was observed (only for the towns of metropolitan France)
- **Author**: name of the author of the picture,
- **YearInCLEF**: ImageCLEF2011, ImageCLEF2012, ImageCLEF2013, PlantCLEF2014 when the image was integrated in the benchmark

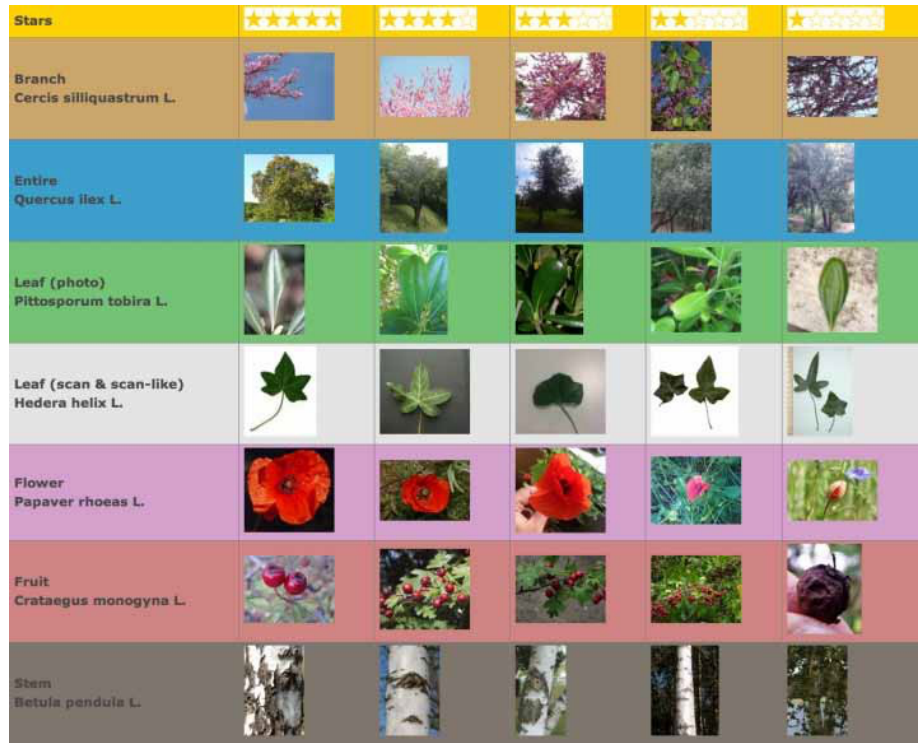

**Fig. 1.** Examples of PlantCLEF pictures with decreasing averaged users ratings for the different types of views

- **IndividualPlantId2013**: the plant observation ID used last year during the ImageCLEF2013 plant task,
- **ImageID2013**: the image id.jpg used in 2013.

## 3   Task Description

The task was evaluated as a plant species retrieval task based on multi-image plant observations queries. The goal was to retrieve the correct plant species among the top results of a ranked list of species returned by the evaluated system. Contrary to previous plant identification benchmarks, queries are not defined as single images but as *plant observations*, meaning a set of one to several images depicting the same individual plant, observed by the same person, the same day, with the same device. Each image of a query observation is associated with a single view type (entire plant, branch, leaf, fruit, flower, stem or leaf scan) and with contextual meta-data (data, location, author). Each participating group was allowed to submit up to 4 runs built from different methods. Any human assistance in the processing of the test queries has therefore to be signaled in the submitted runs meta-data.

In practice, the whole PlantCLEF dataset was split in two parts, one for training (and/or indexing) and one for testing. All observations with pictures used in the previous plant identification tasks were directly integrated in the training dataset. Then for the new observations and pictures, in order to guarantee that most of the time each species contained more images in the training dataset than in the test dataset, we used a constrained random rule for putting with priority observations with more distinct organs and views in the training dataset. The test set was built by choosing 1/2 of the observations of each species with this constrained random rule, whereas the remaining observations were kept in the reference training set. Thus, 1/3 of the pictures are in the test dataset (see Table 1 for more detailed stats). The xml files containing the meta-data of the *query* images were purged so as to erase the taxon names (the ground truth) and the image quality ratings (that would not be available at query stage in a real-world mobile application). Meta-data of the observations in the training set are kept unaltered.

Table 1: Detailed content of the LifeCLEF 2014 Plant Task dataset.

|  | Total | Branch | Entire | Flower | Fruit | Leaf | LeafScan | Stem |
|---|---|---|---|---|---|---|---|---|
| **Observations** | | | | | | | | |
| Train | **11341** | 1173 | 3969 | 5328 | 1608 | 5196 | 2386 | 1823 |
| Test | **8163** | 574 | 2488 | 3529 | 744 | 1522 | 294 | 567 |
| All | **19504** | 1747 | 6457 | 8857 | 2352 | 6718 | 2680 | 1823 |
| **Images** | | | | | | | | |
| Train | **47816** | 1987 | 6356 | 13165 | 3753 | 7754 | 11335 | 3466 |
| Test | **13146** | 731 | 2983 | 4559 | 1184 | 2058 | 696 | 935 |
| All | **60962** | 2718 | 9339 | 17724 | 4937 | 9812 | 12031 | 4401 |

The metric used to evaluate the submitted runs is a score related to the rank of the correct species in the returned list. Each query observation is attributed with a score between 0 and 1 reflecting equal to the inverse of the rank of the correct species (equal to 1 if the correct species is the top-1 decreasing quickly while the rank of the correct species increases). An average score is then computed across all plant observation queries. A simple mean on all plant observation test would however introduce some bias. Indeed, we remind that the PlantCLEF dataset was built in a collaborative manner. So that few contributors might have provided much more observations and pictures than many other contributors who provided few. Since we want to evaluate the ability of a system to provide the correct answers to all users, we rather measure the mean of the average classification rate per author. Finally, our primary metric was defined as the following average classification score S:

$$S = \frac{1}{U}\sum_{u=1}^{U}\frac{1}{P_u}\sum_{p=1}^{P_u} s_{u,p} \qquad (1)$$

where $U$ : number of users (who have at least one image in the test data), $P_u$: number of individual plants observed by the u-th user, $s_{u,p}$: the score between 1 and 0 equals to the inverse of the rank of the correct species (for the p-th plant observed by the u-th user).

A secondary metric was used to evaluate complementary (but not mandatory) runs providing species prediction at the image level. Each test image is attributed with a score between 0 and 1: of 1 if the 1st returned species is correct and decrease quickly while the rank of the correct species increases. An average score is then be computed on all test images. Following the same motivations expressed above, a simple mean on all test images would however introduce some bias. Some authors sometimes provided many pictures of the same individual plant (to enrich training data with less efforts). Since we want to evaluate the ability of a system to provide the correct answer based on a single plant observation, we also have to average the classification rate on each individual plant. Finally, our secondary metric is defined as the following average classification score S:

$$S = \frac{1}{U}\sum_{u=1}^{U}\frac{1}{P_u}\sum_{p=1}^{P_u}\frac{1}{N_{u,p}}\sum_{n=1}^{N_{u,p}} s_{u,p,n} \qquad (2)$$

where $U$ is the number of users, $P_u$ the number of individual plants observed by the $u$-th user, $N_{u,p}$ the number of pictures of the $p$-th plant observation of the $u$-th user, $s_{u,p,n}$ is the score between 1 and 0 equals to the inverse of the rank of the correct species.

## 4 Participants and methods

74 research groups worldwide registered to the plant task (31 of them being exclusively registered to the plant task). Among this large raw audience, 10 research groups did cross the finish line by submitting runs (from 1 to 4 depending on the teams). 6 teams submitted 14 complementary runs on images.

Participants were mainly academics, specialized in computer vision, machine learning and multimedia information retrieval. We list below the participants and give a brief overview of the techniques they used in their runs. We remind here that LifeCLEF benchmark is a system-oriented evaluation and not a deep or fine evaluation of the underlying algorithms. Readers interested by the scientific and technical details of any of these methods should refer to the LifeCLEF 2014 working notes of each participant (referenced below):

**BME TMIT (3 runs), [29], Hungary.** These participants used Gaussian Mixture Model (GMM) based Fisher vector (FV) representation from a dense-SIFT features extraction. Principal Component Analysis (PCA) is first used on the SIFT vectors in order to reduce the dimension from 128 to 80. Then a Fisher Vector representation based on a concise codebook of 256 visual words is used for embedding the PCA-SIFT descriptors in a single high level representation for each image. The chosen classifier was the C-Support Vector Classification (C-SVC) algorithm with Radial Basis Function kernel. The two hyperparameters (C from C-SVC and from RBF kernel) were optimized by a grid search with two-dimensional grid. The algorithm was trained with the training image set, and then validated on the validation set, while the hyperparameters were different in each iteration. In order to obtain a final species list for all test images from a same observation, since the C-SVC classifier calculates continuous reliability value for each class at each image, they used a combined classifier using a weighted average of reliability values.

**FINKI (3 runs) [7], Macedonia.** For the LeafScan category these participants used the multiscale triangular shape descriptor [24]. For the other pictures, opponent SIFT were extracted around 20 000 points of interest obtained using Harris-Laplace detector. In addition, a rhomboid-shaped mask was applied to the input image to minimize the effect of the cluttered background and to reduce the number of points as in [4]. Then an approximate k-means (AKM) algorithm is used for clustering these descriptors and for producing a large number of visual words (approximately 200K). Then, they used for each test image a Bag-of-visual-Word representation and used a classical TF-IDF measure in order to compute a training image list. Finally, in order to combine the result lists from several test images belonging to a same plant observation, two fusion operators as experimented: *Min rank* (run 2), a "*Probability fusion*" (run 3). The combination of the two approaches (run 1) gave their best results. Then a 1-nn rule is used at the end for producing a list of ranked species.

**I3S (2 runs) [15], France.** These participants used a Bag-of-Word framework starting from SIFT and Opponent Color SIFT features extracted from 1000 points localised with the SIFT detector. Several visual dictionaries were constructed with a K-means clustering algorithms: K=4000 words for the LeafScan category, K=2000 for the Leaf, and K=500 for the other types of views. Then, 3500 (7 image categories x 500 species) SVM binary classifiers were trained in order to give for each test image a list of species with a decreasing normalized score of confidence. Pictures from a same test plant observation are gathered according to two rules: sum (Run 1) and max (Run 2) of confidence normalized scores.

**IBM AU (4 runs) [6], Australia.** These participants tested and combined distinct approaches. Run1 uses a deep Convolutional Neural Network. Their CNN has around 60 million parameters and is composed of 5 convolutional lay- ers, some of which are followed by max-pooling layers, and three fully-connected layers with a final softmax layer. They followed the pipeline in [21], but they restricted the node number of the fully connection network to 2048 as this number is far more enough to model the plant images. Run 2 used a Gaussian Mixture Model based Fisher Kernel approach. First, they extracted dense SIFT and Color Moments (CMs) in images and each local feature was reduced to a 64-dim after using PCA. Then for each type of features, a GMM model with 512 components is estimated for producing two Fisher Vector representations by image. Then, they averaged the output from linear trained SVMs classifiers, one for each type of features. In Run 3 & 4 they combined the CNN and the FV approaches with an empirical rule. Run 4 is like the run 3 but with a segmentation preprocessing step on images in order to find better regions on interest to analyse (only on Flower, Fruit, Leaf, LeafScan and Stem).

**IV-Processing (1 run) [8], Tunisia.** These participants embedded several types of features (the outputs of the harris detector, a haar wavelet decomposition, a RGB color histogram) into a single binary code after a Principal Component Analysis step. Then the hamming distance is used in order to compare images from the training dataset with image query.

**MIRACL (3 runs) [19], Tunisia.** This team tested visual and textual approaches. For the visual part they used a combination of standard global descriptors: Color Layout Descriptor (CLD), Edge Orientation Histogram (EOH) and a Scalable Color Descriptor (SCD). Then, they attempted to use the contextual content in the associated XML documents for each image with textual and structural representations.

**PlantNet (4 runs) [13], France.** These participants used for all categories a large scale matching approach, and some shape descriptors in the specific case of LeafScan (Directional Fragment Histogram and standard shape parameters). A geometrically constrained multi-scale & multi-orientation Harris-Laplace detector is used in order detect around 100-150 points mostly located at the center of the pictures. Then, they extracted numerous local features: SURF, Fourier2D, rotation invariant Local Binary Patterns, Edge Orientation Histogram, weighted RGB, weighted-LUV and HSV histograms. After preliminary evaluations, each type of view had its own subsets of types of local features. These local features are hashed, indexed and searched in separate index with the Random Maximum Margin Hashing approach (one for each type of view and for each type of feature). Then, a hierarchical late fusion scheme is applied in order to combine the image response lists of the different modalities: first from the different types of local features, then from the multiple-images from a same category, and finally from all the categories in order to obtain a final list related to one plant observation. Different fusion algorithms are experimented in order to combine the information at each level: a weighted probabilities approach (run 1), and the BordaMNZ count (run 2,3,4) and IprMNZ count (run 4 only for LeafScan and Fruit) inspired from the voting theory. The final species list is produced thanks to an adaptive k-nn rule (k being related to plant observations, not images).

**QUT (1 run) [28], Australia.** These participants used (*Overfeat*) a Convolutional Neural Network pre-trained with the generic ImageNet dataset previously used to perform general object classification. They used two layers in this network in order to obtain two sets of visual features: the Layer 17 gave a set of 3072-dim vectors, and Layer 19 gave a set of 4096-dim vectors. Then, a extremely Randomized Trees Classifier is used in order to output a probability distribution over the 500 species, one for each feature. The probability distributions are then averaged in order to compute a single probability distribution for a test image. Finally, probability distributions from several pictures from a same test observation are added in order to obtain the final list of ranked species. Note that this kind of approach was not really allowed since the Overfeat features are pre-trained with some external resources.

**Sabanki-Okan (2 runs) [30], Turkey.** These participants used distinct approaches, depending to the category of picture. For the LeafScan category, an automatic segmentation was performed using edge preserving morphological simplification by means of area attribute filters, followed by an adaptive threshold. Then, a variety of shape and texture features are extracted (the same than the ones used during the previous 2012-13 campaigns): Circular Covariance Hist. (CCH), Rotation Invariant Point Triplets (RIT), Orientation Hist. (OH), Color Auto-correlogram, etc. For the Flower, Fruit and Entire categories, they used a Bag of Visual Word approach: they extracted some dense-SIFT features and used a K-Means in order to obtain the visual dictionary of 1200 words and produce BoW representations for each image. For the Stem category, they used the same global descriptors on texture and color used for the LeafScan category (CCH, OH, RIT) with an additional Morphological Covariance descriptor. Finally in each system they used SVM classifiers for predicting a list of ranked species. In the specific case of Branch and Leaf categories, they used a Convolutional Neural Network approach. The CNN employed contains 8 layers where the output of the last fully connected softmax layer produces a distribution over the species.

**SZTE (4 runs) [26], Hungary.** The SZTE focused their work on the LeafScan category: after a first Otsu segmentation, they extracted a Vein density description, various shape parameters (area/perimeter, perimeter/diameter, diameter/perpendicular-diameter) and the cumulative histogram representation of Multiscale Triangular shape descriptors successfully evaluated in last year plant identification task [18], [24]. Then a Random Forest classifier is used for predicting species. For the other categories, they used a Color-Gradient Histogram CGH on pictures. The test images are then compared with the training images and k-nn classifier gave a ranked species list. Finally an heuristic rule is proposed for combining pictures from a same test plant observation by allowing a priority to the LeafScan images.

Table 2 attempts to summarize the methods used at different stages (feature, classification,...) in order to highlight the main choices of participants.

## 5 Results

### 5.1 Main task

The following graphic 2 and table 3 show the scores obtained on the main task on plant observation queries.

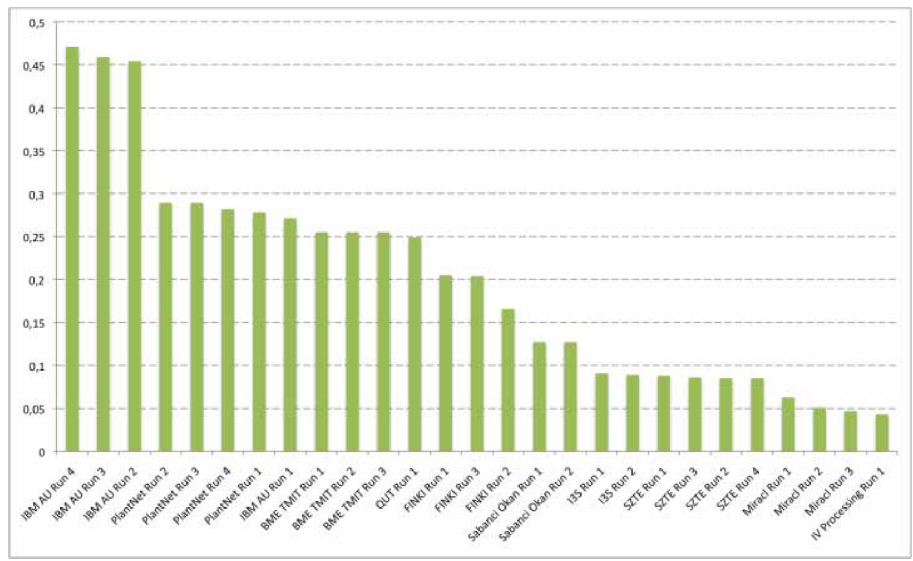

**Fig. 2.** Official results of the LifeCLEF 2014 Plant Identification Task.

The best results are indisputably obtained by the three last runs of the IBM AU team. This results confirms that the Fisher Vector encoding is currently the state-of-art as a generic approach in most of the problems in computer vision. Convolutional Neural Networks, which is an another well-know recent state-of-art technique for object recognition, have not performed as well here in this problem of plant identification as we can see in the first run of IBM AU or the Sabanki-Okan runs. The main reason, as discussed in the working note of IBM AU team [6], is that deep models usually require much training data to learn their millions of parameters and avoid overfitting (e.g. up to 1000 images per class within ImagNet). To solve this issue, deep neural networks are usually pre-trained on generalist classification tasks before being fine-tuned on the targeted task. But as using external training data was not authorized in PlantCLEF

Table2: Approaches used by participants. Column "Obs." indicates if participants avoid to split images from a same plant Observation during evaluation on training dataset.

| Team | Preprocessing | Features | Features representation | Classification/ Similarity search | Image Combination | Meta-data | Obs. |
|---|---|---|---|---|---|---|---|
| BME TMIT | | dense-SIFT + PCA (reducing 128 to 80 dim.) | GMM FV based (256x80) | C-SVC with RBF kernel | weighted average of reliability values | × | × |
| FINKI | Otsu segmentation for LeafScan | *LeafScan*: Multiple triangular representations *Photos*: Harris-like corner detection + opponent SIFT | BoW (200k Words) | 1-nn rule | late hierarchical fusion schema: combination of list images results (Min Rank and/or Probability fusion). | × | |
| I3S | × | SIFT and Opponent-Color SIFT | BoWs | binary SVMs | weighted-sum or max score of confidence | × | |
| IBM AU | segmentation of region of interest (run 4) | CNN (runs 1,3,4) dense-SIFT and CMs (runs 2,3,4) | GMM based FV (run 2,3,4) | Logistic Regression on CNN SVM on FVs | | × | ✓ |
| IV-Processing | × | Harris corner detection, Haar wavelet transform, RGB hist. | hamming distance | | | × | |
| MIRACL | | CLD, EOH, SCD | | euclidian distance | | ✓ | |
| PlantNet | Otsu-like segmentation for LeafScan | *LeafScan*: DFH and shape parameters *All*: Harris-like corner detection + SURF, ri-LBP, w-RGB, w-Luv, HSV, Fourier2D, EOH, Hough | RMMH | adaptive k-nn rule | late hierarchical fusion schema: BordaMNZ, IrpMNZ rules, weighted probabilities combination | × | ✓ |
| QUT | downscale to 231 pixels | *Overfeat* pre-trained CNN: Layer 17¿a set of 3072-dim vectors, Layer 19¿a set of 4096-dim vectors | | Extremely Randomized Trees | image level: average of probability distributions from the prediction on each feature, observation level: likelihood sum | × | |
| Sabanki-Okan | Auto segmentation for LeafScan | *Scan*: CCH, RIT, OH, CAC, etc *Flower, Fruit, Entire*: dense-SIFT *Branch, Leaf*: CNN *Stem*: CCH, RIT, OH, CAC, MC | Fruit,Flower,Entire: BoW Leaf,Branch: CNN | LeafScan, Fruit, Flower, Entire, Stem: SVMs Leaf,Branch: output of CNN | | date on flower & fruit | ✓ |
| SZTE | Otsu segmentation for LeafScan | LeafScan: Vein density, shape parameters Photos: CGH | | LeafScan: Random Forest Photos: knn | heuristic | × | |

Table 3: Results of the LifeCLEF 2014 Plant Identification Task. Column "Keywords" attempts to give the main idea of the method used in each run. Fisher Vector (FV) and Bags Of Visual Word (BoW) representations involve SVM classifiers.

| Run name | Key-words | Metadata | Score |
| --- | --- | --- | --- |
| IBM AU Run 4 | FV+CNN+Segm. | × | 0,471 |
| IBM AU Run 3 | FV+CNN | × | 0,459 |
| IBM AU Run 2 | FV | × | 0,454 |
| PlantNet Run 2 | Matching RMMH | × | 0,289 |
| PlantNet Run 3 | Matching RMMH | × | 0,289 |
| PlantNet Run 4 | Matching RMMH | × | 0,282 |
| PlantNet Run 1 | Matching RMMH | × | 0,278 |
| IBM AU Run 1 | CNN | × | 0,271 |
| BME TMIT Run 1 | FV | × | 0,255 |
| BME TMIT Run 2 | FV | × | 0,255 |
| BME TMIT Run 3 | FV | × | 0,255 |
| QUT Run 1 | CNN feat. | × | 0,249 |
| FINKI Run 1 | BoW | × | 0,205 |
| FINKI Run 3 | BoW | × | 0,204 |
| FINKI Run 2 | BoW | × | 0,166 |
| Sabanci-Okan Run 1 | BoW, CNN | date | 0,127 |
| Sabanci-Okan Run 2 | BoW, CNN | date | 0,127 |
| I3S Run 1 | BoW | × | 0,091 |
| I3S Run 2 | BoW | × | 0,089 |
| SZTE Run 1 | Global mixed | × | 0,088 |
| SZTE Run 3 | Global mixed | × | 0,086 |
| SZTE Run 2 | Global mixed | √ | 0,085 |
| SZTE Run 4 | Global mixed | √ | 0,085 |
| Miracl Run 1 | Global | √ | 0,063 |
| Miracl Run 2 | Global, Textual, Structural | | 0,051 |
| Miracl Run 3 | Global, Textual, Structural | × | 0,047 |
| IV Processing Run 1 | Local/Global mixed | | 0,043 |

2014, this approach could not be evaluated by the participants. Allowing such approaches in next campaigns might be possible but is a tricky problem as we need to guaranty that none of the images of test set could be found somewhere on the web.

Despite the supremacy of IBM fisher vectors runs, it is surprising to see that the performances of BME TMIT runs, which are based on a very close training model, reached much lower performances. It demonstrates that different implementations and parameters tuning can bring very different performances (e.g. 512x60 fisher vectors dimensions for IBM AU vs. 258x80 for BME TMIT, IBM AU used additionnal Color Moments descriptors while BME TMIT used only SIFT).Morever, like it was demonstrated during previous ImageCLEF Plant Identification Task campaigns, teams who split the training data according to the observation id during their preliminary evaluations on validation sets, seem have to take benefit of it, avoiding certainly overfitting problems like for the IBM AU, PlantNet teams for instance. BME TMIT did not mentioned that and may be were in this case, explaining also the difference of performances with the IBM AU runs.

Another outcome is that the second best performing method from PlantNet was already among the best performing methods in previous plant identification challenges [4] although LifeCLEF dataset is much bigger and somehow more complex because of the social dimension of the data. This demonstrates the genericity and stability of the underlying matching method and features.

This year few teams attempted to explore the metadata. The date was exploited in the Sabanki-Okan runs, only on flowers or fruits, but we don't have a point of comparison in order to see if the use of this information was useful or not. Miracl team attempted to combine the whole textual and structural informations contained in the xml files, but it has been showed to degrade the performances of their pure visual approach. Note that for the first year, after three years of unsuccessful attempts during the previous ImageCLEF Plant Identification Tasks, none of the teams explored the locality and GPS information.

### 5.2 Complementary results on images

6 teams submitted 14 complementary runs on images. The following graphic 3 and table 4 below present the scores obtained on the complementary run files focusing on images. Thanks to the participants who produced these not mandatory run files. In order to evaluate the benefit of the combination of the test images from the same observation, the graphic compares the pairs of run files on images and on observations assumed to have been produced with the same method (it is not the case for the BME TMIT team). Basically, for each method, we can observe a substantial improvement by combining the different views from a same plant observation. It is a good news, since this is the current practice of botanists, who most of the time can't identify a species with only one picture on only one organ. However, we can say that the improvement are not so much high: we guess that there is a room of improvement here, basically with more images and may be with new methods of fusions dealing with this specific problem of multi-image and multi-organ problem.

Table 4: Comparison between results on identification by images with results on identification by observation

| Run name | Score Image | Score Observation |
|---|---|---|
| IBM AU Run 4 | 0,456 | 0,471 |
| IBM AU Run 3 | 0,446 | 0,459 |
| IBM AU Run 2 | 0,438 | 0,454 |
| PlantNet Run 2 | 0,28 | 0,289 |
| PlantNet Run 4 | 0,276 | 0,282 |
| PlantNet Run 1 | 0,271 | 0,278 |
| IBM AU Run 1 | 0,263 | 0,271 |
| FINKI Run 1 | 0,205 | 0,205 |
| FINKI Run 3 | 0,204 | 0,204 |
| FINKI Run 2 | 0,166 | 0,166 |
| Sabanci Okan Run1 | 0,123 | 0,127 |
| Sabanci Okan Run2 | 0,123 | 0,127 |
| BME TMIT Run 1 | 0,086 | 0,255 |
| I3S Run 1 | 0,043 | 0,091 |

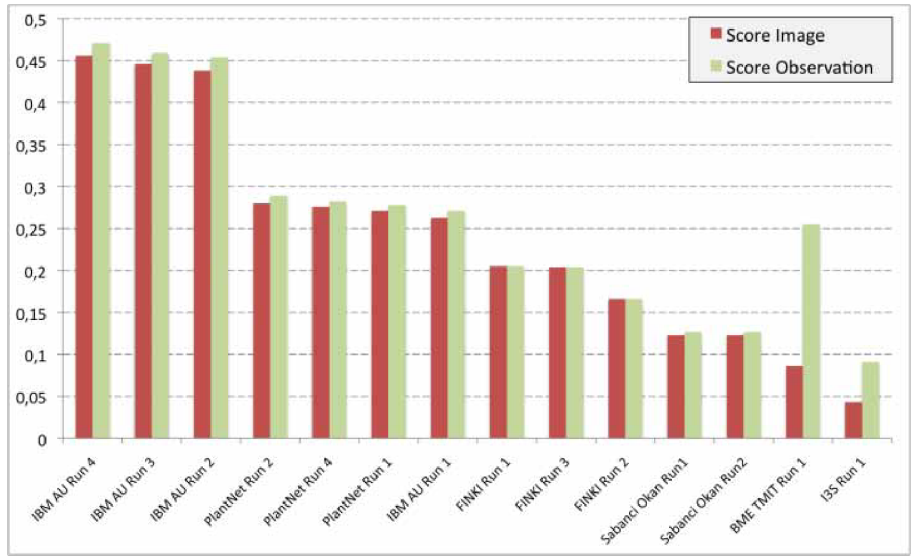

**Fig. 3.** Comparison of the methods: before and after combining the prediction for each image from a same plant observation.

### 5.3 Complementary results on images detailed by organs

The following table 5 and graphic 4 below show the detailed scores obtained for each type of organs. Remember that we use a specific metric weighted by authors and plants, and not by sub-categories, explaining why the score on images not detailed is not the mean of the 7 scores of these sub-categories. Like during the previous Plant Identification Task, the LeafScan and the Flower categories obtained the best results, while it is not easy to rank the other organ by difficulty (maybe Fruit, Entire, Leaf, Stem, Branch). Results obtained on the Branch category by the three last runs of IBM AU outperformed completely the other approaches, while more shape dedicated approaches reduce the difference with this generic approach on LeafScan (PlantNet, Sabanki-Okan & Finki). Interestingly, the pure CNN approach in IBM AU Run 1 obtained rather good results on Flower, the organ where there is a lot of data (as much as in the LeafScan category in terms of number of observations), confirming the potential of the CNN approach with more data.

Table 5: Detailed by organ results on complementary run files on test images

| Run name | Branch | Entire | Flower | Fruit | Leaf | Leaf Scan | Stem |
|---|---|---|---|---|---|---|---|
| IBM AU Run 4 | 0,292 | 0,333 | 0,585 | 0,339 | 0,318 | 0,64 | 0,269 |
| IBM AU Run 3 | 0,298 | 0,34 | 0,57 | 0,326 | 0,304 | 0,614 | 0,267 |
| IBM AU Run 2 | 0,294 | 0,335 | 0,555 | 0,317 | 0,3 | 0,612 | 0,267 |
| PlantNet Run 2 | 0,112 | 0,181 | 0,376 | 0,22 | 0,164 | 0,453 | 0,156 |
| PlantNet Run 4 | 0,112 | 0,167 | 0,366 | 0,197 | 0,165 | 0,541 | 0,152 |
| PlantNet Run 1 | 0,112 | 0,168 | 0,366 | 0,197 | 0,165 | 0,449 | 0,133 |
| IBM AU Run 1 | 0,103 | 0,193 | 0,389 | 0,161 | 0,103 | 0,278 | 0,138 |
| FINKI Run 1 | 0,088 | 0,117 | 0,255 | 0,177 | 0,16 | 0,4 | 0,157 |
| FINKI Run 3 | 0,088 | 0,117 | 0,255 | 0,177 | 0,162 | 0,36 | 0,159 |
| FINKI Run 2 | 0,108 | 0,099 | 0,187 | 0,16 | 0,14 | 0,399 | 0,18 |
| Sabanci Okan Run1 | 0,007 | 0,077 | 0,149 | 0,118 | 0,066 | 0,449 | 0,089 |
| Sabanci Okan Run2 | 0,007 | 0,077 | 0,149 | 0,118 | 0,066 | 0,449 | 0,089 |
| BME TMIT Run 1 | 0,052 | 0,06 | 0,115 | 0,07 | 0,019 | 0,119 | 0,072 |
| I3S Run 1 | 0,041 | 0,023 | 0,04 | 0,04 | 0,035 | 0,089 | 0,086 |

## 6 Conclusion

This paper presented the overview and the results of LifeCLEF 2014 plant identification testbed following the three previous campains within ImageCLEF. The number of participants was around 10 groups showing an interest in applying multimedia search technologies to environmental challenges. This year the challenge climb one step by considering multiple type of view and organs of plants while the number of species increased from 250 to 500. Results are encouraging by scaling state-of-the-art plant recognition technologies to a real-world application with thousands and thousands of species might still be a difficult task. With the emergence of more and more plant identification apps [22] [9], [1], [27] and the ecological urgency to build real-world and effective identification tools, we believe that the results and working notes produced during the task will be of high interest for the computer vision and machine learning community. A possible evolution for a new plant identification task in 2015 is to extend the task to all French flora which is estimated to around 5000 species.

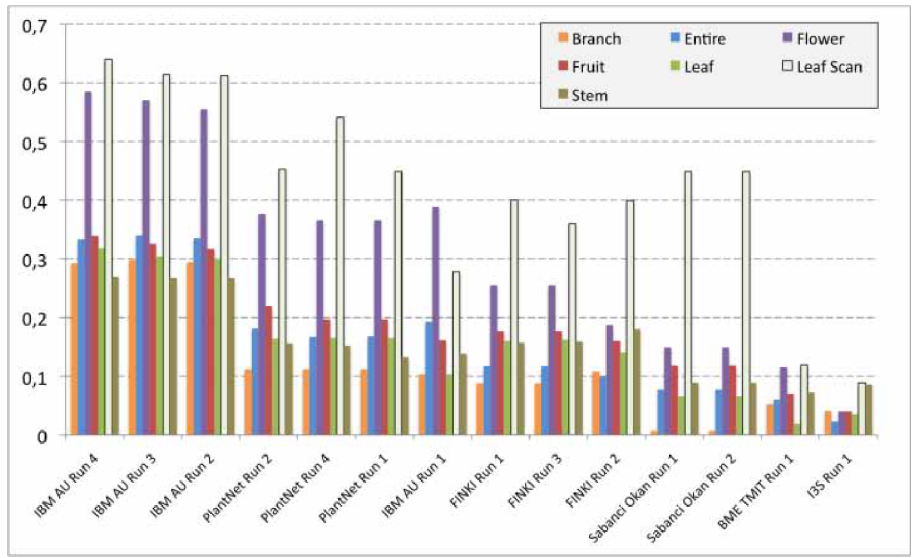

**Fig. 4.** Results detailed for each type of image category.

## References

1. Angelova, A., Zhu, S., Lin, Y., Wong, J., Shpecht, C.: Development and deployment of a large-scale flower recognition mobile app. In: NEC Labs America Technical Report (December 2012), http://www.nec-labs.com/research/information/infoAM_website/pdfs/MobileFlora.pdf
2. Aptoula, E., Yanikoglu, B.: Morphological features for leaf based plant recognition. In: Proc. IEEE Int. Conf. Image Process., Melbourne, Australia. p. 7 (2013)
3. Backes, A.R., Casanova, D., Bruno, O.M.: Plant leaf identification based on volumetric fractal dimension. International Journal of Pattern Recognition and Artificial Intelligence 23(6), 1145–1160 (2009)

4. Bakic, V., Mouine, S., Ouertani-Litayem, S., Verroust-Blondet, A., Yahiaoui, I., Go¨eau, H., Joly, A.: Inria's participation at imageclef 2013 plant identification task. In: CLEF (Online Working Notes/Labs/Workshop) 2013. Valencia, Espagne (2013)
5. Cerutti, G., Tougne, L., Vacavant, A., Coquin, D.: A parametric active polygon for leaf segmentation and shape estimation. In: International Symposium on Visual Computing. pp. 202–213 (2011)
6. Chen, Q., Abedini, M., Garnavi, R., Liang, X.: Ibm research australia at lifeclef2014: Plant identification task. In: Working notes of CLEF 2014 conference (2014)
7. Dimitrovski, I., Madjarov, G., Lameski, P., Kocev, D.: Maestra at lifeclef 2014 plant task: Plant identification using visual data. In: Working notes of CLEF 2014 conference (2014)
8. Fakhfakh, S., Akrout, B., Tmar, M., Mahdi, W.: A visual search of multimedia documents in lifeclef 2014. In: Working notes of CLEF 2014 conference (2014)
9. Go¨eau, H., Bonnet, P., Joly, A., Affouard, A., Bakic, V., Barbe, J., Dufour, S., Selmi, S., Yahiaoui, I., Vignau, C., Barth´el´emy, D., Boujemaa, N.: Plantnet mobile 2014: Android port and new features. In: Proceedings of International Conference on Multimedia Retrieval (2014)
10. Go¨eau, H., Bonnet, P., Joly, A., Boujemaa, N., Barth´el´emy, D., Molino, J.F., Birnbaum, P., Mouysset, E., Picard, M.: The ImageCLEF 2011 plant images classification task. In: CLEF working notes (2011)
11. Go¨eau, H., Bonnet, P., Joly, A., Yahiaoui, I., Barthelemy, D., Boujemaa, N., Molino, J.F.: The imageclef 2012 plant identification task. In: CLEF working notes (2012)
12. Go¨eau, H., Joly, A., Selmi, S., Bonnet, P., Mouysset, E., Joyeux, L., Molino, J.F., Birnbaum, P., Bathelemy, D., Boujemaa, N.: Visual-based plant species identification from crowdsourced data. In: ACM conference on Multimedia (2011)
13. Go¨eau, H., Joly, A., Yahiaoui, I., Bakic, V., Anne, V.B.: Pl@ntnet's participation at lifeclef 2014 plant identification task. In: Working notes of CLEF 2014 conference (2014)
14. Hazra, A., Deb, K., Kundu, S., Hazra, P., et al.: Shape oriented feature selection for tomato plant identification. International Journal of Computer Applications Technology and Research 2(4), 449–meta (2013)
15. Issolah, M., Lingrand, D., Precioso, F.: Plant species recognition using bag-of-word with svm classifier in the context of the lifeclef challenge. In: Working notes of CLEF 2014 conference (2014)
16. Joly, A., Goeau, H., Bonnet, P., Bakic, V., Barbe, J., Selmi, S., Yahiaoui, I., Carr´e, J., Mouysset, E., Molino, J.F., Boujemaa, N., Barth´el´emy, D.: Interactive plant identification based on social image data. Ecological Informatics (2013), http://www.sciencedirect.com/science/article/pii/S157495411300071X
17. Joly, A., Go¨eau, H., Bonnet, P., Bakic, V., Molino, J.F., Barth´el´emy, D., Boujemaa, N.: The imageclef plant identification task 2013. In: International workshop on Multimedia analysis for ecological data. Barcelone, Espagne (Oct 2013), http://hal.inria.fr/hal-00908934
18. Joly, A., Go¨eau, H., Bonnet, P., Bakic, V., Molino, J.F., Barth´el´emy, D., Boujemaa, N., et al.: The imageclef plant identification task 2013. In: International workshop on Multimedia analysis for ecological data (2013)
19. Karamti, H., Fakhfakh, S., Tmar, M., Gargouri, F.: Miracl at lifeclef 2014: Multi-organ observation for plant identification. In: Working notes of CLEF 2014 conference (2014)

20. Kebapci, H., Yanikoglu, B., Unal, G.: Plant image retrieval using color, shape and texture features. The Computer Journal 54(9), 1475–1490 (2011)
21. Krizhevsky, A., Sutskever, I., Hinton, G.E.: Imagenet classification with deep convolutional neural networks. In: Advances in Neural Information Processing Systems (2012)
22. Kumar, N., Belhumeur, P.N., Biswas, A., Jacobs, D.W., Kress, W.J., Lopez, I.C., Soares, J.V.B.: Leafsnap: A computer vision system for automatic plant species identification. In: European Conference on Computer Vision. pp. 502–516 (2012)
23. Mouine, S., Yahiaoui, I., Verroust-Blondet, A.: Advanced shape context for plant species identification using leaf image retrieval. In: ACM International Conference on Multimedia Retrieval. pp. 49:1–49:8 (2012)
24. Mouine, S., Yahiaoui, I., Verroust-Blondet, A.: A shape-based approach for leaf classification using multiscale triangular representation. In: ICMR '13 - 3rd ACM International Conference on Multimedia Retrieval (2013)
25. Nilsback, M.E., Zisserman, A.: Automated flower classification over a large number of classes. In: Indian Conference on Computer Vision, Graphics and Image Processing. pp. 722–729 (2008)
26. Paczolay, D., Bánhalmi, A., Nyúl, L., Bilicki, V., Sárosi, Á.: Wlab of university of szeged at lifeclef 2014 plant identification task. In: Working notes of CLEF 2014 conference (2014)
27. Reves: Folia (Nov 2012), https://itunes.apple.com/fr/app/folia/id547650203
28. Sunderhauf, N., McCool, C., Upcroft, B., Tristan, P.: Fine-grained plant classification using convolutional neural networks for feature extraction. In: Working notes of CLEF 2014 conference (2014)
29. Szúcs, G., Dávid, P., Lovas, D.: Viewpoints combined classification method in image-based plant identification task. In: Working notes of CLEF 2014 conference (2014)
30. Yanikoglu, B., S. Tolga, Y., Tirkaz, C., FuenCaglartes, E.: Sabanci-okan system at lifeclef 2014 plant identification competition. In: Working notes of CLEF 2014 conference (2014)